\pdfoutput=1
\documentclass[11pt]{article}

\usepackage[]{EMNLP2022}

\usepackage{fancyhdr}

\fancypagestyle{firstpage}{
\fancyhf{}
  \fancyfoot[C]{UM-IoS workshop, EMNLP 2022, Abu-Dhabi, UAE.}
}

\usepackage{times}
\usepackage{latexsym}

\usepackage[T1]{fontenc}
\usepackage[utf8]{inputenc}

\usepackage{microtype}

\usepackage{inconsolata}

\usepackage{subfigure}

\usepackage{graphicx}
\usepackage{amsmath}
\usepackage{amssymb}

\usepackage{titlesec}

\title{Visual Grounding of Inter-lingual Word-Embeddings}

\author{ Wafaa Mohammed 
\hspace{1pt}
Hassan Shahmohammadi
\hspace{1pt} 
Hendrik P. A. Lensch 
\hspace{1pt} 
R. Harald Baayen \\
         \\ University of Tübingen \\ 
         wmohammed@aimsammi.org, \\
         \{hassan.shahmohammadi, hendrik.lensch, harald.baayen\}@uni-tuebingen.de
         }
\begin{document}
\maketitle
\begin{abstract}

Visual grounding of Language aims at enriching textual representations of language with multiple sources of visual knowledge such as images and videos. Although visual grounding is an area of intense research,  inter-lingual aspects of visual grounding have not received much attention. The present study investigates the inter-lingual visual grounding of word embeddings. We propose an implicit alignment technique between the two spaces of vision and language  in which inter-lingual textual information interact in order to enrich pre-trained textual word embeddings. We focus on three languages in our experiments, namely, English, Arabic, and German. We obtained visually grounded vector representations for these languages and studied whether visual grounding on one or multiple languages improved the performance of embeddings on word similarity and categorization benchmarks. Our experiments suggest that inter-lingual knowledge improves the performance of grounded embeddings in similar languages such as German and English. However, inter-lingual grounding of German or English with Arabic led to a slight degradation in performance on word similarity benchmarks. On the other hand, we observed an opposite trend on categorization benchmarks where Arabic had the most improvement on English. In the discussion section, several reasons for those findings are laid out. We hope that our experiments provide a baseline for further research on inter-lingual visual grounding.

% \hlnote{improvment on what exactly?}\hsnote{it should be clear now}
\end{abstract}

\section{Introduction}
\thispagestyle{firstpage}
\label{intro}

% talk about linking language and vision
Distributional Semantic Models (DSMs) have long been used to capture words' meaning. They estimate semantic representations from co-occurrences of words in text corpora.   Even though embeddings are the dominant method for large scale data, from a psychological and cognitive point of view, distributional models suffer from the problem referred to as the \textit{Symbol Grounding Problem} \citep{harnad1990symbol}: %which is the problem of models where  RHB
the meaning of a symbol (word) is entirely accounted for in terms of other symbols without any links to the outside world.
% Cognitive scientists have argued that DSMs can never recover the full meaning of words just from text because words' meanings are grounded in our perceptual and motor systems \citep{de2008levels}. In cognitive science, grounding is defined as `` the process of establishing what mutual information is required for successful communication between two interlocutors '' \citep{chandu2021grounding}.
In the context of natural language processing (NLP), grounding is defined as `` the process of linking the symbolic representation of language (e.g., words) into the rich perceptual knowledge of the outside world '' \citep{shahmohammadi2021learning}. 
% adding the theoritcal study
Moreover, \citep{huang2021makes} have proved that multi-modal learning outperforms uni-modal learning as it has access to a better quality latent space representation.

Many studies have addressed grounding of language in vision, typically focusing on grounding for English \citep{bruni2014multimodal,shahmohammadi2022language}. As a consequence, inter-lingual visual grounding is still poorly understood. This study investigates whether monolingual textual embeddings benefit from the knowledge of other languages in the process of visual grounding.  We extend a state-of-the-art model for monolingual visual grounding \citep{shahmohammadi2022language} by considering different combinations of three languages, namely, English, German, and Arabic. Using various word categorization benchmarks, our experiments show that the three languages profitably exchange inter-lingual knowledge across a simple linear vector space. To the best of our knowledge, we are the first to investigate the problem of visual grounding of inter-lingual word embeddings. Overall, our contributions are as follows: 

% \begin{itemize}
%     \item We propose a simple extension of a state-of-the-art visual grounding model to integrate three different languages.
%     \item We obtain zero-shot visually grounded embeddings in three languages.
%     \item Using various benchmarks, we reveal how visual grounding changes textual vector space across languages and show that inter-lingual knowledge transfers to downstream tasks.
% \end{itemize}
\noindent
a) We propose a simple extension of a state-of-the-art visual grounding model to integrate three different languages. b) We obtain zero-shot visually grounded embeddings in three languages. c) Using various benchmarks, we reveal how visual grounding changes textual vector space across languages and show that inter-lingual knowledge transfers to downstream tasks.

Our paper is structured as follows: Section~\ref{Related_work} briefly highlights the related works. Section~\ref{interlingual_grounding} introduces our problem of interest. In Section~\ref{model_architecture} our proposed model is elaborated. Implementation details are covered in Section~\ref{implementation_details}. The results are presented in Section~\ref{results}, with further discussion in section~\ref{discussion}. In Section~\ref{conclusion}, we conclude our research, and finally, we point out the limitations and future directions of our work.

% \textcolor{red}{the following are missing from the introduction}
% \begin{itemize}
%     \item not that many works on inter-lingual visual grounding, we (are the first) or we wanna do this
%     \item general overview of our model
%     \item general and interesting finding from our paper
%     \item optional: how the paper is structured
% \end{itemize}

% add paper structure

\section{Related Work}
\label{Related_work}
% \hsnote{I think we need to cover more previous works}
% \hsnote{write a short intro + create two sections: visual grounding, and inter-lingual visual grounding, something like: }

% \begin{itemize}
%     \item There have been many works attempting to enrich word embeddings with vision. most of them focus on a single language and multi-lingual visual grounding is limited.
%     \item \textbf{single language grounding:} mention more previous works concisely. you dont need to explain everything
%     \item \textbf{interlingual visual grounidng:} explain the task briefly, cite previous works if any, mention our works falls into this category.
    
% \end{itemize}
% talk about previous work on grounding language into vision, specially on English.

There have been many studies on language grounding in vision most of which focus on monolingual visual grounding. There have been also other works on cross-modal and cross-lingual representations tailored for specific downstream applications. 

\textbf{Monolingual grounding:} The study of \citet{bruni2014multimodal} was one of the first studies to obtain visually grounded embeddings by simple fusion such as applying SVD on the concatenation of word and image vectors. \citep{kiros2018illustrative} adopted a similar fusion approach using gating mechanisms. \citep{silberer2014learning} and \citep{hasegawa2017incorporating} encoded the two modalities as vectors of attributes and combine them using autoencoders.
% \citep{collell2017imagined} optimized for a regression task of predicting an image vector given its corresponding textual representation. 
\citep{kurach2017better} and \citep{shahmohammadi2022language} adopted a simple approach where textual embeddings are directly optimized to match image representations. They propose a grounding framework that depends on the alignment of textual and visual features.

\textbf{Cross-modal cross-lingual representations:} In the multilingual setting, the focus has largely been on cross-modal downstream tasks. \citep{burns2020learning} proposed a  scalable multilingual aligned language representation using masked cross-language modelling objective. \citep{ni2021m3p} proposed a multilingual multimodal model that combines different languages and different modalities into a shared space via multitask pre-training. Similarly, \citep{zhou2021uc2} introduced a machine translation augmented model for cross-modal cross-lingual learning by introducing multi-modal losses. 
% \citep{aggarwal2020towards} used metric learning to train a model on English image-sentence pairs, then they extend the model to multilinguality in a zero-shot fashion during inference. 
\citep{mohammadshahi2019aligning} trained a multilingual multimodal model by optimizing the alignment between languages for image-description retrieval task. 

The present study is inspired by both directions explored in the literature on visual grounding and multi-lingual representations.  We propose a straight forward alignment technique informing textual representations about the visual space while also making use of inter-lingual features. We generate visually grounded inter-lingual word embeddings and evaluate their performance on similarity and categorization benchmarks.

A new direction of research that has been published in parallel with this paper is the work of \citep{chen2022pali}. Their model, PaLI (Pathways Language and Image model), employs scaling of joint vision and language pre-training. They make use of the largest transformers to date to train the model. They were able to achieve state-of-the-art in multiple vision and language tasks such as captioning, visual question answering, and scene-text understanding. 

% show that this paper is an extension to hassan's work in a multilingual setting

\section{Inter-lingual Visual Grounding}
\label{interlingual_grounding}
Multilingual-language models hold great promise for the development of embeddings for under-resourced languages \citep{armengol2021multilingual}. The central idea in this line of research is that different languages bring different perspectives (e.g., cultural information and grammar) which can inform each other, resulting in a richer model that has a better understanding of words' meanings in any specific language. Moreover, since typical visual scenes are thought to produce similar information across different languages, integrating visual knowledge (e.g., images) into a multilingual model can contribute to  obtaining a better quality grounded embedding space.
%that is closer to the real world. 

%wafaa: the model architecture is in the next section!!
% In this section, we describe in detail our proposed model for inter-lingual visual grounding.
%hassan: not sure if this is necessary
% As a result, this makes language technologies such as voice assistants, chatbots, etc. accessible for everyone in the world regardless of the language they speak.

\section{Model Architecture}
\label{model_architecture}
% model description
\label{arch1}

% image is to be modified
\begin{figure*}
    \centering
    \includegraphics[width=17cm, height=6cm]{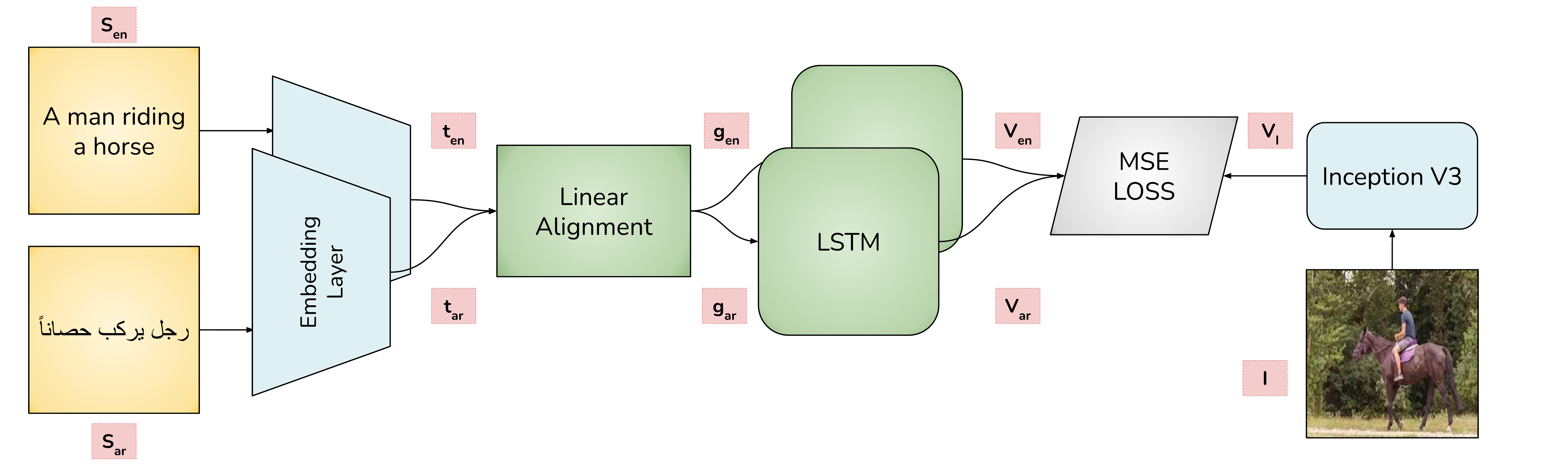}
    \caption{Model Architecture. sentences are first tokenized. Individual tokens are passed, one by one, to a pre-trained embedding layer, followed by a linear alignment that transfers the embeddings into the grounded space.  Grounded vectors are encoded into a single vector by an LSTM encoder. The output of the LSTM is then optimized against the image vector generated via a pre-trained CNN model. Layers in blue are frozen during training. } % add caption
    \label{fig:model1}
\end{figure*}

% Our model maps a textual description of an image into its corresponding image representation. The model makes use of a linear alignment to preserve most of the textual knowledge in the word embeddings, allowing only subtle modifications by the error received from the image. The model is trained using multilingual image captioning data. These datasets comprise images along with their captions in multiple languages. The model is given the task to match, for a given image,  the captions in two (or three) languages to that image in such a way that language-specific features are preserved, and not overwhelmed by inter-lingual features, and image features. 

Our model maps a textual description of an image into its corresponding image representation. It makes use of a linear alignment to preserve most of the textual knowledge in the word embeddings, allowing only subtle modifications by the error received from the image. It is trained using multilingual image captioning data. The model is given the task to match, for a given image,  the multilingual captions to that image in such a way that language-specific features are preserved, and not overwhelmed by inter-lingual features, and image features. 

Our model maps two (or three) languages to the grounded space using a shared linear alignment. For instance, figure \ref{fig:model1} introduces the model for the combination of English and Arabic languages. Let $D$ be the dataset consisting of triple samples of \((I,S_{en},S_{ar})\in D\). Here $I$ refers to an image, $S_{en}$ and $S_{ar}$ denote matching captions of $I$ in English and Arabic respectively. As shown in Figure~\ref{fig:model1}, the two captions are passed through a pre-trained embedding layer (GloVe) \citep{pennington2014glove} to obtain their textual representations $t_{en}, t_{ar}$ which are then mapped to a visually grounded space through a linear transformation. We refer to this linear transformation as the alignment layer. The alignment layer is used to extract grounded embeddings after training. During training, grounded word vectors of each caption are encoded as a single vector using an LSTM layer as follows:

\( V_{en} = LSTM_{en} (g_{en}, c_0, h_0 | \hspace{1pt} \theta), \) 

\( V_{ar} = LSTM_{ar} (g_{ar}, c_0, h_0 | \hspace{1pt} \theta) \)

\noindent
where, $g_{en}, g_{ar}$ denote the grounded word vectors of the English and Arabic captions respectively. $c_0$, $h_0$ and $\theta$ represent the initial cell state, initial hidden state, and the trainable parameters of the LSTM. The parameters of the linear alignment and the LSTM layer are optimized to match the sentence representations in both languages to the same image vector $V_{I}$ as follows:
 
%\(\hat{\Theta}_{en}= \underset{\Theta}{argmin} \hspace{1pt}\frac{1}{|D|} \sum_{t=1}^{n} (V_{I}^t - V_{en}^t)^2,\) \\

%\(\hat{\Theta}_{ar}= \underset{\Theta}{argmin} \hspace{1pt}\frac{1}{|D|} \sum_{t=1}^{n} (V_{I}^t - V_{ar}^t)^2,\) \\

\(  \mathcal{L}_{en}(\theta_{en}) = \hspace{1pt}\frac{1}{|D|} \sum_{t=1}^{n} (V_{I}^t - V_{en}^t)^2,\)

\( \mathcal{L}_{ar}(\theta_{ar}) = \hspace{1pt}\frac{1}{|D|} \sum_{t=1}^{n} (V_{I}^t - V_{ar}^t)^2,\) 

\noindent
where $\theta_{en}$ and $\theta_{ar}$ indicate the learning parameters for each language. The image vector $V_{I}^*$  is generated using a pre-trained CNN model. The overall loss is simply the sum of the two losses: $\mathcal{L}_{all}(\Theta) = \mathcal{L}_{en}(\theta_{en}) +  \mathcal{L}_{ar}(\theta_{ar})$

\noindent
In this equation, $\Theta$ represents all the network's learning parameters. After training, we generate grounded word embedding using the alignment layer. A given textual word embedding $w_t \in \mathbb{R}^d$ is passed through the trained alignment, after which its grounded version is extracted from the alignment layer: $g_t \in \mathbb{R}^c$ as $g_t = w_t.M$, where $M$ denotes the trained alignment layer.

\section{Implementation details}
\label{implementation_details}
We used the Microsoft COCO 2017 dataset  \citep{lin2014microsoft} for our experiments. This  dataset consists of 123,287 images with 5 captions each. It is split into 118k training images and 5k validation images. We experimented with three languages for the captions: English, Arabic, and German. The original dataset provided by Microsoft contains the English captions. We obtained the German captions from \citep{biswas2021improving}, who translated the English COCO captions using the Fairseq neural machine translator, and the Arabic captions from \citep{MuhammadHashim}, who generated the captions using Google's advanced cloud translation API. 
% Three experiments implemented visual grounding for each of the three languages, three experiments implemented visual grounding for one language, while fusing in a second language, and one experiment visually grounded all three languages jointly.
For the Arabic version of COCO, we only had available to us translations of the captions for 82k samples, which we split into 77k samples for training and 5k samples for validation, and this is the set of images that we use for models that included Arabic. 
For fair comparisons, we also investigated model performance for English and German using the same 82k images. For all the experiments, we used TensorFlow as a development framework
. The training environment is similar to the one used by 
%and followed the implementation of
\citet{shahmohammadi2022language}.  We used a batch size of 256 image-caption pairs. 
%which included 256 images along with one of its captions. 
We trained for 20 epochs with 5 epochs as early stopping tolerance, using the NAdam optimizer \citep{dozat2016incorporating} with a learning rate of 0.001. The image vectors were obtained using  pre-trained vectors from Inception-V3 \citep{szegedy2016rethinking}, which are based on ImageNet \citep{deng2009imagenet}. For pre-trained textual embeddings we used GloVe embeddings \citep{pennington2014glove}. The vocabulary considered for training English comprised the 10k most frequent words. For German and Arabic, which have much richer inflectional systems compared to English, we took into account the 30k most frequent words.  We set the dimension of grounded word embeddings to 1024 ($g_t \in \mathbb{R}^{1024}$), and matched the size of the LSTM's output to that of the image vectors (both to 2048). Both the embedding layer and the pre-trained CNN were frozen during training.

\section{Results}
\label{results}
In this section, we explain our evaluation criteria and report the results of our experiments. We use various word similarity/relatedness and word categorization benchmarks and provide both quantitative and qualitative results. 

\subsection{Qualitative Evaluation}
\begin{figure*}
    \centering
    \includegraphics[width=.95\textwidth, height=7cm]{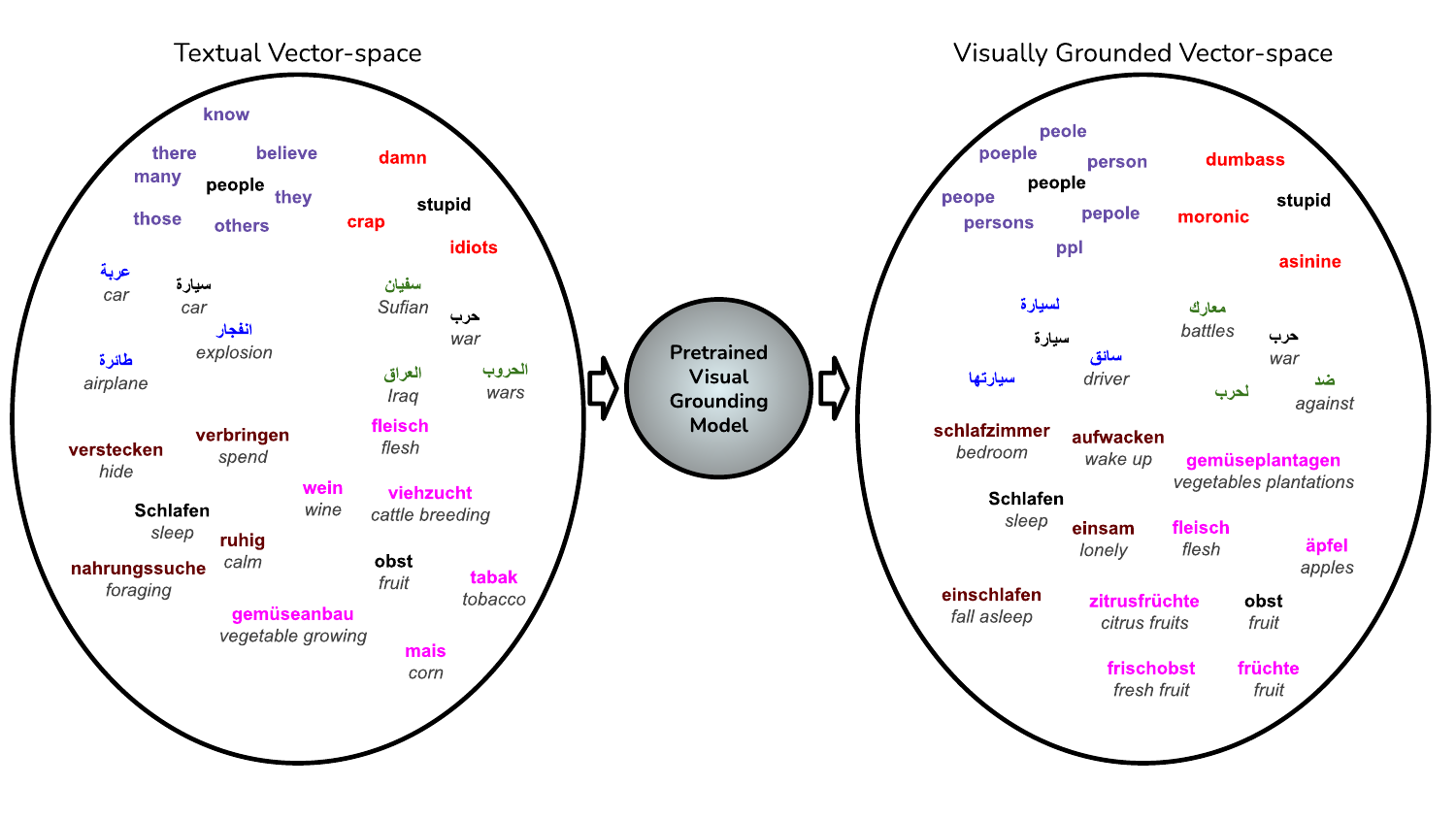}
    \caption{Comparisons of the textual and grounded vector spaces for English, German, and Arabic. For each query word (in black), out of the 10 nearest neighbours, the neighbours unique to each space are displayed. Visual grounding better captures a word's meaning and reduces the dependency on just co-occurrence statistics.} % add caption
    \label{fig:quality}
\end{figure*}

Figure \ref{fig:quality} shows the difference between the nearest neighbours of words from the three languages in the textual and grounded spaces (using the grounding setup with separate grounding of each individual language).  The representations in the grounded space are semantically much more precise, and are much less dependent on simple co-occurrence statistics.  Our algorithm for visual grounding thus contributes to taking a step forward in solving the symbol grounding problem. For example, the word \textit{car} in Arabic has its nearest neighbours as \textit{airplane} and \textit{explosion} in the textual space, while in the grounded space, the neighbours are different declensions of the word \textit{car}.

\subsection{Word Similarity/ Relatedness Evaluation}

Following \citep{bruni2014multimodal, shahmohammadi2022language}, we evaluated our visually grounded word embeddings using similarity/relatedness benchmarks. The task is to estimate the similarity/relatedness score of a pair of words using the Spearman correlation as evaluation metric.
Relatedness is a measure of the extent to which two words are associated with each other, e.g.\ (pen, paper). Similarity 
%is a subset of relatedness that
quantifies how alike two concepts are based on their location within an is-a hierarchy (e.g., car, automobile). Some benchmarks differentiate between the two while others consider them similar when scoring pairs of words.
% For example, the pair (student, professor) has a score of $2$ (out of $10$) in the SimLex999 benchmark which distinguishes between similarity and relatedness, while the score of the same pair is $6.81$ in the WordSim353 benchmark, which does not distinguish between relatedness and similarity. 

Tables \ref{english}, \ref{german}, \ref{arabic} summarize the results of visually grounded embeddings on similarity/relatedness benchmarks for English, German, and Arabic. For English, we experimented with six similarity/relatedness benchmarks: WordSim353 \citep{finkelstein2001placing}, MEN \citep{bruni2014multimodal}, RW \citep{Luong-etal:conll13:morpho}, MTurk \citep{radinsky2011word}, simVerb \citep{gerz2016simverb}, and SimLex999 \citep{hill2015simlex}. For German, evaluations are based on the Multilingual versions of WrdSim353 and simLex999 \citep{leviant2015separated}. For Arabic,  similarity was evaluated using four benchmarks: Almarsoomi \citep{almarsoomi2013awss}, MC30 \citep{hassan2009cross}, Saif40 \citep{saif2014evaluating}, and WordSim \citep{hassan2009cross}.

Across the three languages, visual grounding yields embeddings that perform substantially better than embeddings that are based on text only. It is noteworthy that the grounded embeddings achieved superior results  on all the similarity benchmarks, for all three languages. 

For both English and German, adding German and English respectively as a second language to the model leads to a further improvement in performance on the benchmark tasks.   Adding Arabic as a second language along with English or German, however, led to a reduction in accuracy.  The experiments evaluating Arabic word embeddings revealed that fusing in English or German did not improve performance on the Arabic benchmarks.  Furthermore, experiments implementing visual grounding for three languages jointly did not provide further accuracy.

%Moreover, it can also be noticed that the combination of two languages is better than grounding a single language when the languages are similar to each other (using similar alphabets and sharing some words across the two languages) which is the case for English and German. However, when the languages are highly dissimilar (different alphabet), such as Arabic and English or Arabic and German, the combination of the two languages seems to reduce the quality of the grounded embeddings compared to grounding each language separately. The combination of the three languages did not improve much over two languages.  %%% 

The same findings can also be observed even when varying the size of the training and validation data.
%These conclusions do not depend on the size of the training and validation data.  
For example, 
%It can also be noticed that the similarity of the languages is more significant than the training dataset size in the quality of the generated embeddings. For example, 
for the same set of 82k images, adding German embeddings to English embeddings led to an improvement on benchmark tasks, whereas adding Arabic embeddings did not.
%
%more than Arabic on the grounded English embeddings. 
%We hypothesize that inter-lingual features can be beneficial for all the combinations of languages, since they add the different perspectives and cultures of the languages when describing the same visual scene. However, a simple linear alignment does not seem to capture the optimal inter-lingual features for dissimilar languages. \\
In the discussion section, we provide a detailed discussion of why Arabic embeddings do not provide further precision for English or German grounded embeddings.

\begin{table*}
\centering
\begin{tabular}{llllllll}
\hline
\textbf{} & \textbf{WSim} & \textbf{MEN} & \textbf{RW} & \textbf{MTurk} & \textbf{SimVerb} & \textbf{SimLex} & \textbf{Mean}\\
\hline
Textual & 73.8 & 80.5 & 45.5 & 71.5 & 28.3 & 40.8 & 56.7 \\
Grounded EN & 77.7 & \textbf{84.8} & 51.9 & 73.3 & \textbf{38.02} & \textbf{52.2} & 62.9 \\
Grounded EN (82k) & 76.03 & 84.5 & 50.3 & 72.7 & 34.9 & 48.6 & 61.2 \\
Grounded EN + DE & \textbf{79.2} & \textbf{84.8} & \textbf{52.3} & 74.1 & 36.6 & 51.03 & \textbf{63} \\
Grounded EN + DE (82k) & 75.3 & 84.3 & 50.8 & \textbf{74.2} & 34.5 & 49.1 & 61.4 \\
Grounded EN + AR & 76.9 & 84.7 & 50.3 & 73.1 & 34.3 & 48.3 & 61.3 \\
Grounded EN + DE + AR & 76.7 & 84.3 & 51.1 & 73.9 & 33.3 & 48.04 & 61.2 \\

\hline
\end{tabular}
\caption{\label{english}
Performance of textual and grounded English embeddings on similarity/relatedness benchmarks. Results include different combinations of the three languages, English (EN), German (DE), and Arabic (AR). Inter-lingual grounding in English and German outperforms both the textual and monolingual grounded embeddings.
}
\end{table*}

\begin{table*}
\centering
\begin{tabular}{llll}
\hline
\textbf{} & \textbf{WSim} & \textbf{SimLex} & \textbf{Mean}\\
\hline
Textual & 46.6 & 30.9 & 38.8 \\
Grounded DE & 56.2 & 36.9 & 46.6 \\
Grounded DE (82k) & 56.3 & 35.8 & 46.1 \\
Grounded DE + EN & \textbf{57.02} & \textbf{37.2} & \textbf{47.1} \\
Grounded DE + AR & 55.5 & 33.2 & 44.3 \\
Grounded DE + EN (82k) & 56.6 & 35.1 & 45.9 \\
Grounded DE + EN + AR & 54.1 & 33.2 & 43.7 \\

\hline
\end{tabular}
\caption{\label{german}
Performance of textual and grounded German embeddings on similarity/relatedness benchmarks. Results include different combinations of German embeddings with two other languages: English (EN), and Arabic (AR). Grounding in both German and English outperforms all other monolingual groundings. 
}
\end{table*}

\begin{table*}
\centering
\begin{tabular}{llllll}
\hline
\textbf{} & \textbf{WSim} & \textbf{Almarsoomi} &  \textbf{MC30} & \textbf{Saif40} & \textbf{Mean}\\
\hline
Textual & 30.7 & 65.9 & 49.9 & 71.8 & 54.6 \\
Grounded AR & \textbf{41.9} & 72.8 & \textbf{59.2} & 80.6 & \textbf{63.6} \\
Grounded AR + EN & 39.7 &  72.8 &  56.9 & \textbf{83.2} & 63.2 \\
Grounded AR + DE & 36.9 & \textbf{75.2} & 52.6 & 77.05 & 60.4 \\
Grounded AR + EN + DE & 39.6  &  73.9 & 56.2 & 75.5 & 61.3  \\

\hline
\end{tabular}
\caption{\label{arabic}
Performance of textual and grounded Arabic embeddings on similarity/relatedness benchmarks. Results include different combinations of Arabic embeddings with two other languages: English (EN), and German (DE).
% Grounding in both German and English outperforms all other monolingual groundings. 
}
\end{table*}

\begin{table*}
\centering
\begin{tabular}{llllllll}
\hline
\textbf{} & \textbf{Battig} & \textbf{AP} & \textbf{BLESS} & \textbf{ESSLLI-a} & \textbf{ESSLLI-b} & \textbf{ESSLLI-c} & \textbf{Mean}\\
\hline
Textual & 45.4 & 60.4 & \textbf{87.5} & 75.0 & 75.0 & 62.2 & 67.6 \\
Grounded EN  & 47.03 & 60.7 & 80.5 & 75.0 & 75.0 & \textbf{64.4} & 67.1 \\
Grounded EN + DE  & 48.6 & 62.4 & 87 & \textbf{84.1} & 77.5 & 60.0 & \textbf{69.9} \\
Grounded EN + FA  & 47.1 & 64.4 & 85.5 & 81.8 & \textbf{80.0} & \textbf{64.4} & \textbf{70.5} \\
Grounded EN + AR  & \textbf{49.8} & \textbf{64.9} & 79.5 & \textbf{84.1} & 75.0 & \textbf{64.4} & 69.6 \\
Grounded EN + DE + AR  & 47.5 & 64.7 & 85.5 & 75.0 & 75.0 & 62.2 & 68.3 \\
Grounded EN + DE (82k)   & 47.1 & 65.9 & 81.5 & 84.1 & 77.5 & 55.6 & 68.6 \\

\hline
\end{tabular}
\caption{\label{english_category}
Performance of textual and grounded English embeddings on Categorization benchmarks. Results include different combinations of the three languages, English (EN), German (DE), Arabic (AR), and Persian (FA). 
}
\end{table*}

%\hlnote{use boldface for marking best-performing configurations in all tables.} done!
% insert image

\subsection{Word Categorization Evaluation}
\label{cat}
We also evaluated our embeddings on six categorization benchmarks: Battig \citep{battig1969category}, AP \citep{almuhareb2005concept}, BLESS \citep{baroni2011we}, and three tasks published at
%the European Summer School in Logic, Language and Information 
\citep{esslli}, \citep{essllia}, which focuses on grouping concrete nouns into semantic categories; \citep{essllib}, which tests computational models for their ability to discriminate between abstract and concrete nouns; and \citep{essllic}, which groups verbs into semantic categories.

The concept-categorization task requires clustering a set of nouns expressing basic-level concepts into gold standard categories. To evaluate on this task, clustering is performed using a k-means clustering algorithm \citep{likas2003global}.  Performance is evaluated using a purity score between the truth and predicted cluster labels.
Results are presented in Table~\ref{english_category}. Monolingual grounding did not result in improvements on this benchmark; grounded English embeddings revealed worse performance on BLESS compared to the textual embeddings. However, adding a second language solved this problem.  Incorporation of both German and Arabic embeddings resulted in improved performance of the English embeddings on all benchmarks.  However, combining the three languages did not give rise to further improvements.  Interestingly,  for the smaller dataset size (82k images), Arabic had a better performance than German, a result that contrasts with those obtained for the similarity benchmarks.
% \noindent

\textbf{More Languages:} We further extended our experiments by using the Persian language. For this aim, we translated the COCO captions using google translate API\footnote{https://libraries.io/pypi/googletrans} and made use of a pre-trained GloVe word embeddings model\footnote{https://github.com/taesiri/PersianWordVectors} train on OSCAR \citep{2022arXiv220106642A}. Similar to other languages grounding textual Persian embeddings significantly boosted the result (Spearman's correlation) by more than $10 \%$ (from $36.7$ to $47$) on the SemEval2017 benchmark \citep{camacho2017semeval}. Due to time constraints, we only trained the grounded embeddings from English + Persian and evaluated them on the word categorization benchmarks. As shown in Table~\ref{english_category}, Adding Persian (denoted as FA) results in the best mean performance. 

%Further analysis of those results are presented in section \ref{analysis}.

\begin{figure*}[htb]
    \centering
    \subfigure[Textual English embeddings]{\includegraphics[width=0.47\textwidth]{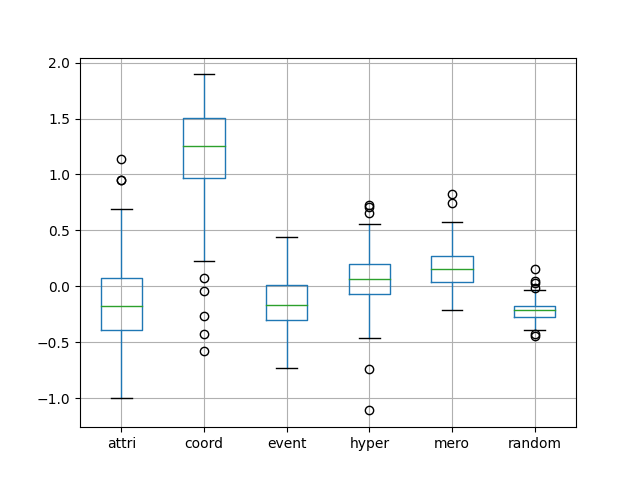}}{}\subfigure[Grounded English embeddings]{\includegraphics[width=0.47\textwidth]{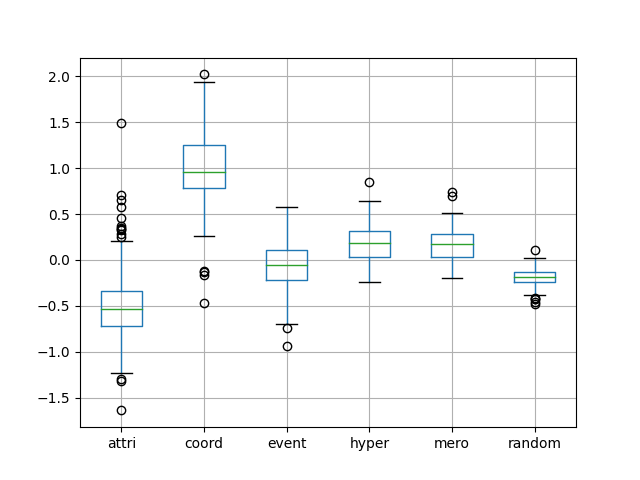}}{}
    \subfigure[Grounded English + German embeddings]{\includegraphics[width=0.47\textwidth]{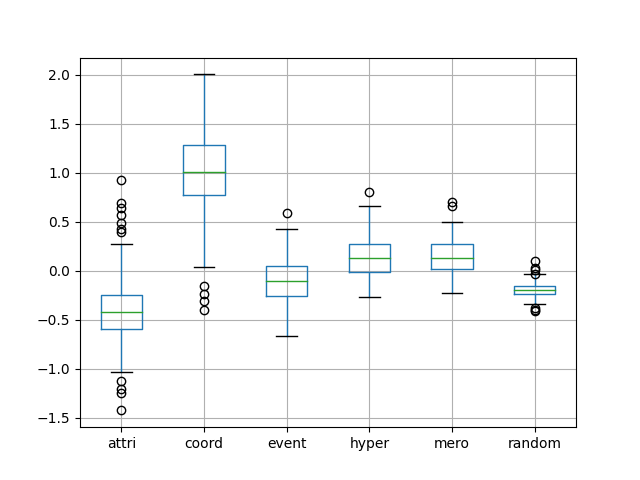}}{}\subfigure[Grounded English + Arabic embeddings]{\includegraphics[width=0.47\textwidth]{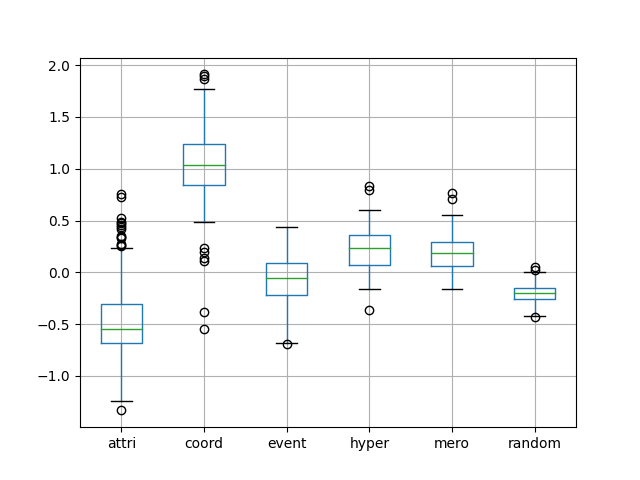}}{}
    \caption{BLESS \citep{baroni2011we} Analyses of textual and grounded English embeddings with the combination of other languages. Visual grounding clearly reduces the variance on \textit{attri} and \textit{coord} categories resulting in more refined clusters and higher word categorization scores.}
    \label{fig:bless}
\end{figure*}

% \section{Analysis}
% \label{analysis}
To further analyze the interaction of visual grounding with multiple languages, we made use of the BLESS \citep{baroni2011we} dataset. This dataset consists of tuples of the format \textit{(concept-relation-relatum)}. For example, \textit{lizard-attri-striped}: the concept \textit{lizard} is linked to the relatum \textit{striped} via the \textit{attribute} relation. BLESS focuses on a set of basic concrete nouns and explicit semantic relations. Additionally, it contains a number of random relatum words that are not semantically related to any of  the concepts. The tasks that come with this dataset it to detect which words are related to a given concept, as well as determining the type of relation involved. The dataset comprises  200 concepts grouped into 17 classes.

BLESS includes 5 types of relations, in-addition to the random relations: \textbf{COORD}: the relatum is a noun that is a co-hyponym (coordinate) of
the concept: \textit{dishwasher-coord-oven}.
\textbf{HYPER}: the relatum is a noun that is a hypernym of the concept: dishwasher-hyper-appliance.
\textbf{MERO} : the relatum is a noun referring to a part/component/organ/member of the concept, or something that the concept contains or is made of: \textit{dishwasher-mero-button}. \textbf{ATTRI} : the relatum is an adjective expressing an attribute of the concept: \textit{dishwasher-attri-full}.
\textbf{EVENT} : the relatum is a verb referring to an action/activity/happening/event the concept is involved in or is performed by/with the concept: \textit{dishwasher-event-use}.

Using our embeddings, we calculated the mean cosine similarity score of each concept to all its relata across all relations. For each of the 200 BLESS concepts, we obtain six cosine similarity scores, one per relation: $C_{ir} = \hspace{1pt}\frac{1}{n} \sum_{j=1}^{n}  cos(C_{i}, Rel_{rj})$ where $C_{ir}$ denotes the mean cosine score of concept $i$ for relation $r$ and $n$ indicates the number of words per relation. The scores are then normalized across each concept as: $C_{ir} = \frac{C_{ir} - \mu_{i}}{\sigma_{i}}$, where $\mu_{i}$ and $\sigma_{i}$ denote the mean and the standard deviation of the scores of $C_i$ across all relations. Figure \ref{fig:bless} presents the distribution of scores per relation across the 200 concepts. While the coarse structures of all the embeddings are relatively similar with respect to the scores (cosine similarity) across relations, the figures reveal interesting properties. For instance, the distributions in both \textit{attri} and \textit{coord} are more compact when visual grounding is applied. That is, the model is more certain about the similarity between the words and hence creates a more refined cluster of words. Another interesting point is the increased mean in the \textit{hyper} category, especially for Arabic,  in line with the results reported in Table~\ref{english_category}.

Moreover, visual grounding lowers the mean score on \textit{coord} category across all languages; this is probably because of the visually different word pairs in \textit{coord} category. For example, (\textit{turtle}, \textit{alligator}) and (\textit{toaster}, \textit{stove}) are not visually similar. Therefore, their word vectors diverge as the result of grounding. These findings are in line with previous findings that visual grounding prioritizes similarity over relatedness \citep{shahmohammadi2021learning}.  Surprising at first sight is that the mean score of \textit{attri} category is lower in all grounding setups. This, however, may be due to the rather different sets of attributes in BLESS and in our image captions.   Many of the attributes used in BLESS rarely occur in image captions, examples are \textit{antarctic}, \textit{amphibious}, \textit{aquatic}, and  \textit{noisy}.

In order to statistically validate these findings, we applied a Gaussian Location-Scale Generalized Additive Mixed Model (GAMM) \citep{Wood:2017}, with word as random-effect factor, and main effects for \textit{embedding type} and \textit{relation} for both mean and variance. This analysis revealed that 
the grounded English embeddings (monolingual grounding) had the highest mean score, followed by the grounded English embeddings generated by integrating English and German, followed closely by the English + Arabic embeddings. Interestingly, compared to the textual embeddings, the variance for grounded embeddings is reduced, and even more reduced for inter-lingual grounded embeddings with Arabic and German. Thus, there seems to be a trade-off between mean and variance.  While monolingual grounding had the highest mean score, inter-lingual grounding helped more in reducing the variance, resulting in more refined clusters of semantically related words.

Comparing the mean of scores with respect to the different relations, with the \textit{random} relation as the baseline, we noticed that 
% compared to this baseline, 
the mean decreases for \textit{attri}, but increases for all other relations, and noticeably so for the \textit{hyper} and \textit{mero} relations. The variance, on the other hand, increases for all relations 
% compared to the random baseline, 
and to the greatest extent for \textit{attr} and \textit{coord}. These statistical results dovetail well with our  previously mentioned conclusions about visually different word pairs in \textit{coord} category and the difference in \textit{attributes} between the BLESS data and our image captions. Overall, the boxplots indicate that inter-lingual visual grounding creates more refined clusters of word vectors in the vector space based on visual clues in the training sets.

\section{Discussion}
\label{discussion}
We proposed an inter-lingual visual grounding model on textual word embeddings. Our model thus far supports the benefit of visual grounding and inter-lingual visual grounding on various word similarity and word categorization benchmarks. Some of the results in Section~\ref{results} however are hard to interpret. In this section, we will discuss possible explanations for the model's behavior on different tasks across different languages.

On the word similarity benchmarks (Tables~\ref{english}, \ref{german}, and \ref{arabic}) we observe that German and English seem to interact more efficiently than Arabic with either.  We believe the slight degradation in performance when adding Arabic might be due to the fact that the Arabic language structure is quite different: much more information is packed into its verbs, and pronouns are used differently and more sparingly. Moreover, its orthography leaves out a lot of phonological information (hardly any vowels), so word embeddings are much more ambiguous relative to English or German. Therefore, the semantic spaces that are constructed are much less similar to that in the two other languages. Apart from the evident differences between Arabic and the other two languages, it is worth mentioning that adding Arabic is far from detrimental. That is, the resulting embeddings (Arabic added) still outperform the textual embeddings significantly. This implies that there exists a linearly aligned common core between the three languages (vector spaces) which as observed in section~\ref{cat}, yielded the lowest variance and more pure vector space. Table~\ref{english_category} further supports these findings. Interestingly, the monolingual grounding of English does not seem to improve the categorization performance, inter-lingual knowledge, on the other hand, results in obvious improvements with respect to the mean score. The opposing impact of adding Arabic on the similarity/relatedness results in contrast to the categorization results indicates the need for further investigation on the evaluation criteria of inter-lingual embeddings.

Furthermore, it is not clear why monolingual visual grounding is more beneficial for word similarity compared to word categorization. We think cultural biases might play a role. For example, our training set (the COCO image dataset) is likely culture-specific, with a strong bias toward the US culture, and our benchmarks are compiled with various purposes across different languages. We, therefore, believe that current evaluation benchmarks only shine light on some facets of the complex interplay of different languages in visual grounding, and further investigation is required for more coherent interpretations.

\section{Conclusions}
\label{conclusion}
% \hsnote{we need to make this longer a bit :D}
% add conclusion
The main purpose of this study is to shed light on the problem of inter-lingual visual grounding. We stated the importance of grounding in language understanding and the cognitive plausibility of text representations. We also suggested a baseline architecture for inter-lingual visual grounding and analyzed the performance of the resulting embeddings on word similarity and categorization benchmarks.

Our findings indicate that inter-lingual features lead to improvements on both similarity and categorization benchmarks with a more significant effect on categorization. Our results on the similarity benchmarks indicate that inter-lingual visual grounding is more beneficial for related languages such as English and German, but can lead to reduced performance when unrelated languages, such as English and Arabic, or German and Arabic, are considered jointly. 
%Furthermore, as far as similarity benchmarks are concerned, results show that inter-lingual visual grounding with our approach is more beneficial for related languages such as English and German, but can lead to reduced performance when unrelated languages, such as English and Arabic, or German and Arabic, are considered jointly. 
On the other hand, Arabic provided the most improvement on categorization benchmarks for grounded English embeddings.

% We provided detailed analysis and discussion for possible reasons of why this might be the case. \\
We hope that these initial steps towards inter-lingual visual grounding inspire further research. Low-resourced languages might benefit from joint processing with high-resourced languages in multi-lingual models but one has to make sure that their unique characteristics are not overwhelmed and masked by datasets acquired in different cultural settings. 
% Moreover, We have reservations on the limit of expectations from multi-lingual models. Languages can be quite different, and assuming true equivalence between them will most likely lead to the unique characteristics of low-resourced languages being overwhelmed and masked by English. We hope that research on multilingual embeddings can help in addressing this issue.

\section*{Limitations}
\label{limitations}
The architecture that we made use of for exploring multi-lingual visual grounding has the limitation that embeddings from different languages, which define high-dimensional spaces that are in all likelihood not congruent, constitute the input for visual grounding.  One direction for future research is to first align the embeddings of different languages. A large  
%We consider the problem of achieving the optimal setup for inter-lingual visual grounding an open question. We leave it to future work to improve on the architecture that we presented and experiment with more languages. Some of the directions for future work include distilling knowledge from a large 
multilingual language model such as XLM \citep{lample2019cross} %to the basic linear alignment architecture, which will 
may help to better capture shared inter-lingual features, while at the same time retaining the linear alignment that restricts the extent to which vision can affect text-based semantics.  Another possibility is to use an unsupervised technique \citep{conneau2017word} to generate cross-lingual embeddings, which can then be used as initializers for our grounding architecture.

% add limitations

% \section*{Ethics Statement}

\section*{Acknowledgements}
% acknowledge the cluster of excellence + the institute that provided the German COCO captions.
%\elnote{Any specific format for the acknowledgement of the cluster of excellence?} \\
This work has been supported by EXC number 2064/1 – Project num-
ber 390727645,  the German Federal Ministry of Education
and Research (BMBF): Tübingen AI Center, FKZ: 01IS18039A, and the
European research council, project WIDE-742545. The authors thank the
International Max Planck Research School for Intelligent Systems (IMPRS-IS)
for supporting Hassan Shahmohammadi. We appreciate the contribution of 
\citep{biswas2021improving} from the German Research Center of Artificial Intelligence (DFKI) who have provided us with the German COCO captions upon request.

\bibliography{anthology,custom}
\bibliographystyle{acl_natbib}

% \appendix

% \section{Ablation Studies}

% \begin{table*}
% \centering
% \begin{tabular}{llllllll}
% \hline
% \textbf{} & \textbf{WSim} & \textbf{MEN} & \textbf{RW} & \textbf{MTurk} & \textbf{SimVerb} & \textbf{SimLex} & \textbf{Mean}\\
% \hline
% Textual & 73.8 & 80.5 & 45.5 & 71.5 & 28.3 & 40.8 & 56.7 \\
% Grounded EN & 76.03 & 84.5 & 50.3 & 72.7 & 34.9 & 48.6 & 61.2 \\
% Grounded EN + DE & 75.3 & 84.3 & 50.8 & 74.2 & 34.5 & 49.1 & 61.4 \\
% Grounded EN + AR & 76.9 & 84.7 & 50.3 & 73.1 & 34.3 & 48.3 & 61.3 \\

% \hline
% \end{tabular}
% \caption{\label{english_ablation}
% Performance of textual and grounded English Word Embeddings on similarity/ relatedness benchmarks using 82k image samples along with their captions for training.
% }
% \end{table*}

% \subsection{Using 82k Image Samples for English}
% \label{sec:appendix_arabic}
% As mentioned in section \ref{implementation_details}, for Arabic captions, we had access to captions of 82k images only compared to the full 123k images in COCO dataset. This study is dedicated to show that it is not the dataset size that affects the performance of the model on Arabic language. For fair comparison, we used the same dataset to train on English and English + German and evaluated the resulting grounded English embeddings on similarity benchmarks. Results are shown in Table~\ref{english_ablation}. It can be seen than the combination of English and German is better than English and Arabic. This proves that the dataset size is not the reason for worse results on Arabic but rather the different language structure and culture.

\end{document}